\documentclass[11pt,letterpaper]{article}
\usepackage{naaclhlt2015}
\usepackage{times}
\usepackage{latexsym}
\usepackage{amsmath}
\usepackage{multirow}
\usepackage{url}
\usepackage{graphicx, graphics} 
\usepackage{array}
\usepackage{tabularx}
\usepackage[usenames,dvipsnames]{color}
\usepackage{fixltx2e}
\usepackage[utf8]{inputenc}
\usepackage[table]{xcolor}
\usepackage[normalem]{ulem}

\title{EliXa: A modular and flexible ABSA platform}
\author{I\~{n}aki San Vicente, Xabier Saralegi\\
	    Elhuyar Foundation\\
	    Osinalde industrialdea 3\\
	    Usurbil, 20170, Spain\\
	    \small{{\tt \{i.sanvicente,x.saralegi\}@elhuyar.com}}
	  \And
	Rodrigo Agerri\\
    IXA NLP Group\\
  	University of the Basque Country (UPV/EHU)\\
  	Donostia-San Sebasti\'{a}n\\
  \small{{\tt rodrigo.agerri@ehu.eus}}}
  
\date{}

\begin{document}
\maketitle
\begin{abstract}
This paper presents a supervised Aspect Based Sentiment Analysis (ABSA) system. Our aim is to develop a modular platform which allows to easily conduct experiments by replacing the modules or adding new features. We obtain the best result in the Opinion Target Extraction (OTE) task (slot 2) using an off-the-shelf sequence labeler. The target polarity classification (slot 3) is addressed by means of a multiclass SVM algorithm which includes lexical based features such as the polarity values obtained from domain and open polarity lexicons. The system obtains accuracies of 0.70 and 0.73 for the restaurant and laptop domain respectively, and performs second best in the out-of-domain hotel, achieving an accuracy of 0.80.
\end{abstract}

\section{Introduction}

Nowadays Sentiment Analysis is proving very useful for tasks such as decision making and market analysis. The ever increasing interest is also shown in the number of related shared tasks organized: TASS \cite{tass2012_overview,tass2013_overview}, SemEval \cite{semeval-2013,semeval-2014_4,semeval-2014_9}, or the SemSA Challenge at ESWC2014\footnote{http://challenges.2014.eswc-conferences.org/index.php/SemSA}. Research has also been evolving towards specific opinion elements such as entities or properties of a certain opinion target, which is also known as ABSA. The Semeval 2015 ABSA shared task aims at covering the most common problems in an ABSA task: detecting the specific topics an opinion refers to (slot1); extracting the opinion targets (slot2), combining the topic and target identification (slot1\&2) and, finally, computing the polarity of the identified word/targets (slot3). Participants were allowed to send one constrained (no external resources allowed) and one unconstrained run for each subtask. We participated in the slot2 and slot3 subtasks.

Our main is to develop an ABSA system to be used in the future for further experimentation. Thus, rather than focusing on tuning the different modules our goal is to develop a platform to facilitate future experimentation. The EliXa system consists of three independent supervised modules based on the IXA pipes tools \cite{agerri2014ixapipes} and Weka \cite{hall_weka_2009}.  Next section describes the external resources used in the unconstrained systems. Sections \ref{sec:slot2} and \ref{sec:slot3} describe the systems developed for each subtask and briefly discuss the obtained results.

\section{External Resources}
\label{sec:extRes}

Several polarity Lexicons and various corpora were used for the unconstrained versions of our systems. To facilitate reproducibility of results, every resource listed here is publicly available.

\subsection{Corpora}
\label{sec:ext_corpora}

For the restaurant domain we used the Yelp Dataset Challenge dataset\footnote{http://www.yelp.com/dataset\_challenge}. Following  \cite{nrcSemeval_2014}, we manually filtered out categories not corresponding to food related businesses (173 out of 720 were finally selected). A total of 997,721 reviews (117.1M tokens) comprise what we henceforth call the \textit{Yelp food corpus} ($C_{Yelp}$). 


For the laptop domain we leveraged a corpus composed of Amazon reviews of electronic devices  \cite{Jo_Amazon_2011}. Although only 17,53\% of the reviews belong to laptop products, early experiments showed the advantage of using the full corpus for both slot 2 and slot 3 subtasks. The \textit{Amazon electronics corpus} ($C_{Amazon}$) consists of 24,259 reviews (4.4M tokens). 
Finally, the English Wikipedia was also used to induce word clusters using word2vec \cite{mikolov2013distributed}.

\subsection{Polarity Lexicons}
\label{sec:ext_Lex}

We generated two types of polarity lexicons to represent polarity in the slot3 subtasks: general purpose and domain specific polarity lexicons. 

A general purpose polarity lexicon $L_{gen}$ was built by combining four well known polarity lexicons: SentiWordnet SWN \cite{baccianella_sentiwordnet_2010}, General Inquirer $GI$ \cite{stone_general_1966}, Opinion Finder $OF$ \cite{wilson_recognizing_2005} and Liu's sentiment lexicon $Liu$ \cite{hu_mining_2004}. When a lemma occurs in several lexicons, its polarity is solved according to the following priority order: $Liu$ $>$ $OF$ $>$ $GI$ $>$ $SWN$. The order was set based on the results of \cite{qwn-ppv_eacl}. All polarity weights were normalized to a $[-1,1]$ interval. Polarity categories were mapped to weights for $GI$ ($neg_+$$\rightarrow$$-0.8$; $neg$$\rightarrow$-0.6; $neg_-$$\rightarrow$-0.2; $pos_-$$\rightarrow$0.2; $pos$$\rightarrow$0.6; $pos_+$$\rightarrow$0.8), $Liu$ and $OF$ ($neg$$\rightarrow$-0.7; $pos$$\rightarrow$0.7 for both). In addition, a restricted lexicon $L_{genres}$ including only the strongest polarity words was derived from $L_{gen}$ by applying a threshold of $\pm$0.6.

\begin{table} [ht]
\begin{center}
{\footnotesize
\setlength{\extrarowheight}{1pt}
\begin{tabular}{|@{\hspace{0.1cm}}p{0.15\columnwidth}|c|c|c|c|}\hline
{\bf Domain} & {\bf Polarity Lexicon} & {\bf Total} \\
\hline
General & $L_{gen}$ &  42,218 \\
\hline
General & $L_{genres}$ &  12,398 \\
\hline
Electronic devices & $L_{Amazon}$ &  4,511 \\
\hline
Food & $L_{Yelp}$ &  4,691 \\
\hline
\end{tabular}
}
\end{center}
\caption{Statistics of the polarity lexicons.}
\label{tab:genLexicons}
\end{table}

Domain specific polarity lexicons $L_{Yelp}$ and $L_{Amazon}$ were automatically extracted from $C_{Yelp}$ and $C_{Amazon}$ reviews corpora. Reviews are rated in a $[1..5]$ interval, being 1 the most negative and 5 the most positive. Using the Log-likelihood ratio (LLR) \cite{Dunning_1993} we obtained the ranking of the words which occur more with negative and positive reviews respectively. We considered reviews with 1 and 2 rating as negative and those with 4 and 5 ratings as positive. LLR scores were normalized to a $[-1,1]$ interval and included in $L_{Yelp}$ and $L_{Amazon}$ lexicons as polarity weights.

\section{Slot2 Subtask: Opinion Target Extraction}
\label{sec:slot2}

The Opinion Target Extraction task (OTE) is addressed as a sequence labeling problem. We use the \emph{ixa-pipe-nerc} Named Entity Recognition system\footnote{https://github.com/ixa-ehu/ixa-pipe-nerc} \cite{agerri2014ixapipes} off-the-shelf to train our OTE models; the system learns supervised models via the Perceptron algorithm as described by \cite{collins_discriminative_2002}. \emph{ixa-pipe-nerc} uses the Apache OpenNLP project implementation of the Perceptron algorithm\footnote{http://opennlp.apache.org/} customized with its own features. Specifically, \emph{ixa-pipe-nerc} implements
basic non-linguistic local features and on top of those a combination of
word class representation features partially inspired by \cite{turian-ratinov-bengio:2010:ACL}. The word representation features use large amounts of unlabeled data. The result is a quite simple but competitive system which obtains the best constrained and unconstrained results and the first and third best overall results.

The local features implemented are: current token and token shape (digits, lowercase, punctuation, etc.) in a 2 range window, previous prediction, beginning of sentence, 4 characters in prefix and suffix, bigrams and trigrams (token and shape). On top of them we induce three types of word representations:

\begin{itemize}
    \item Brown \cite{brown1992class} clusters, taking the 4th, 8th, 12th and
        20th node in the path. We induced 1000 clusters on the Yelp reviews dataset described
        in section \ref{sec:ext_corpora} using the tool implemented by Liang\footnote{https://github.com/percyliang/brown-cluster}.
    \item Clark \cite{clark2003combining} clusters, using the standard
        configuration to induce 200 clusters on the Yelp reviews dataset and 100 clusters on the food portion of the Yelp reviews dataset. 
    \item Word2vec \cite{mikolov2013distributed} clusters, based on K-means applied over the extracted word vectors using the skip-gram algorithm\footnote{https://code.google.com/p/word2vec/}; 400 clusters were induced using the Wikipedia.
\end{itemize}

The implementation of the clustering features looks for the cluster
class of the incoming token in one or more of the clustering lexicons induced
following the three methods listed above. If found, then we add the class as a
feature. The Brown clusters only apply to the token related features, which
are duplicated. We chose the best combination of features using 5-fold cross validation, obtaining 73.03 F1 score
with local features (e.g. constrained mode) and 77.12 adding the word clustering features, namely, in unconstrained mode.
These two configurations were used to process the test set in this task. Table \ref{tab:OTE} lists the official results for the first 4 systems in the task.

\begin{table} [ht]
\begin{center}
{\footnotesize
\setlength{\extrarowheight}{1pt}
\begin{tabular} {|@{\hspace{0.1cm}}l|c|c|c|}
\hline
\textbf{System (type) } & \textbf{ Precision}  & \textbf{Recall}  & \textbf{ F1 score} \\
\hline
Baseline & 55.42 & 43.4 & 48.68 \\
\hline
EliXa (u) & 68.93 & 71.22 & \textbf{70.05} \\	
\hline  
NLANGP (u) & 70.53 & 64.02 & 67.12 \\	
\hline
EliXa (c) & 67.23 & 66.61 & 66.91 \\	
\hline
IHS-RD-Belarus (c)  & 67.58 & 59.23 & 63.13\\	
\hline
\end{tabular}
}
\end{center}
\caption{Results obtained on the slot2 evaluation on restaurant data.}
\label{tab:OTE}
\end{table}

The results show that leveraging unlabeled text is helpful in the OTE task, obtaining an increase of 7 points in recall. It is also worth mentioning that
our constrained system (using non-linguistic local features) performs very closely to the second best overall system  by the NLANGP team (unconstrained).
Finally, we would like to point out to the overall low results in this task (for example, compared to the 2014 edition), due to the very small and difficult training set (e.g., containing many short samples such as ``Tasty Dog!'') which made it extremely hard to learn good models for this task. The OTE models will be made freely available in the \emph{ixa-pipe-nerc} website in time for SemEval 2015.

\section{Slot3 Subtask: Sentiment Polarity}
\label{sec:slot3}

The EliXa system implements a single multiclass SVM classifier. We use the SMO implementation provided by the Weka library \cite{hall_weka_2009}. All the classifiers built over the training data were evaluated via 10-fold cross validation. The complexity parameter was optimized as ($C=1.0$). Many configurations were tested in this experiments, but in the following we only will describe the final setting. 

\subsection{Baseline}
\label{sec:slot3_ngram}

The very first features we introduced in our classifier were token ngrams. Initial experiments showed that lemma ngrams (lgrams) performed better than raw form ngrams. One feature per lgram is added to the vector representation, and lemma frequency is stored. With respect to the ngram size used, we tested up to 4-gram features and improvement was achieved in laptop domain but only when not combined with other features.

\subsection{PoS}
\label{sec:slot3_pos}

PoS tag and lemma information, obtained using the IXA pipes tools \cite{agerri2014ixapipes}, were also included as features. One feature per PoS tag was added again storing the number of occurrences of a tag in the sentence. These features slightly improve over the baseline only in the restaurant domain.

\subsection{Window}
\label{sec:slot3_win}

Given that a sentence may contain multiple opinions, we define a  window span around a given opinion target (5 words before and 5 words after). When the target of an opinion is null the whole sentence is taken as span. Only the restaurant and hotel domains contained gold target annotations so we did not use this feature in the laptop domain. 

\subsection{Polarity Lexicons}
\label{sec:slot3_lexicons}

The positive and negative scores we extracted as features from both general purpose and domain specific lexicons. Both scores are calculated as the sum of every positive/negative score in the corresponding lexicon divided by the number of words in the sentence. Features obtained from the general lexicons provide a slight improvement. $L_{genres}$ is better for restaurant domain, while $L_{gen}$ is better for laptops. Domain specific lexicons $L_{Amazon}$ and $L_{Yelp}$ also help as shown by tables \ref{tab:ablatPolarity_rest} and  \ref{tab:ablatPolarity_lapt}.

\subsection{Word Clusters}
\label{sec:slot3_clusters}

Word2vec clustering features combine best with the rest as shown by table \ref{tab:ablatPolarity_rest}. These features only were useful for the restaurant domain, perhaps due to the small size of the laptops domain data.

\subsection{Feature combinations}
\label{sec:slot3_comb}

Every feature, when used in isolation, only marginally improves the baseline. Some of them, such as the E\&A features (using the gold information from the slot1 subtask) for the laptop domain, only help when combined with others. Best performance is achieved when several features are combined. As shown by tables 4 and 5, improvement over the baseline ranges between 2,8\% and 1,9\%  in the laptop and restaurant domains respectively.

\begin{table} [ht]
\begin{center}
{\footnotesize
\setlength{\extrarowheight}{1pt}
\begin{tabular} {|@{\hspace{0.1cm}}p{0.7\columnwidth}|c|}
\hline
 Classifier & Acc Rest \\ 
\hline
Baseline (organizers) & 78.8 \\ 
\hline
{\bf Baseline} & \\
\hspace{0.5cm}1lgram & 80.11 \\ 
\hspace{0.5cm}2lgram & 79.3 \\ 
\hline
$1lgram+E\&A$ & 79.8 \\ 
\hline
$1lgram(w5)$ & 80.41 \\ 
\hline
$1lgram+PoS$ & 80.59 (c) \\ 
\hline
{\bf Lexicons} & \\
\hspace{0.3cm}$1lgram+L_{gen}$ & 80.6 \\ 
\hspace{0.3cm}$1lgram+L_{genres}$ & 81 \\ 
\hspace{0.3cm}$1lgram+L_{Yelp}$ & 80.9 \\ 
\hline
{\bf Combinations} & \\
\hspace{0.3cm}$1lgram(w5)+w2v(C_{Yelp})+L_{genres}+L_{Yelp}+PoS$ & 82.34 (u)\\ 
\hline
\end{tabular}
}
\end{center}
\caption{Slot3 ablation experiments for restaurants. (c) and (u) refer to constrained and unconstrained tracks.}
\label{tab:ablatPolarity_rest}
\end{table}

\begin{table} [ht]
\begin{center}
{\footnotesize
\setlength{\extrarowheight}{1pt}
\begin{tabular} {|@{\hspace{0.1cm}}p{0.7\columnwidth}|c|}
\hline
 Classifier & Acc Lapt \\ 
\hline
Baseline (organizers) & 78.3\\ 
\hline
{\bf Baseline} & \\
\hspace{0.5cm}1lgram & 79.33\\ 
\hspace{0.5cm}2lgram & 79.7\\ 
\hline
$1lgram+clusters(w2v)$  & 79.23\\ 
\hline
$1lgram+E\&A$ & 79.23\\ 
\hline
$1lgram+PoS$ & 78.88 \\ 
\hline
{\bf Lexicons} & \\
\hspace{0.3cm}$1lgram+L_{gen}$ & 79.2\\ 
\hspace{0.3cm}$1lgram+L_{genres}$ & 79\\ 
\hspace{0.3cm}$1lgram+L_{Amazon}$ & 79.7\\ 
\hline
{\bf Combinations} & \\
\hspace{0.3cm}$1lgram+PoS+E\&A$ & 79.99 (c)\\
\hspace{0.3cm}$2lgram+PoS+E\&A$ & 78.27 \\
\hspace{0.3cm}$1lgram+L_{genres}+L_{Amazon}+PoS+E\&A$ & 80.85 (u)\\ 
\hline
\end{tabular}
}
\end{center}
\caption{Slot3 ablation experiments for laptops; (c) and (u) refer to constrained and unconstrained tracks.}
\label{tab:ablatPolarity_lapt}
\end{table}

\subsection{Results}
\label{sec:slot3_res}

Table \ref{tab:slot3_res} shows the result achieved by our sentiment polarity classifier. Although for both restaurant and laptops domains we obtain results over the baseline both performance are modest. 

In contrast, for the out of domain track, which was evaluated on hotel reviews our system obtains the third highest score. Because of the similarity of the domains, we straightforwardly applied our restaurant domain models. The good results of the constrained system could mean that the feature combination used may be robust across domains. With respect to the unconstrained system, we suspect that such a good performance is achieved due to the fact that word cluster information was very adequate for the hotel domain, because $C_{yelp}$ contains a 10.55\% of hotel reviews. 

\begin{table} [ht]
\begin{center}
{\footnotesize
\setlength{\extrarowheight}{1pt}
\begin{tabular} {|@{\hspace{0.1cm}}l|c|c|c|}
\hline
\textbf{System } & \textbf{ Rest.}  & \textbf{Lapt.}  & \textbf{Hotel} \\
\hline
Baseline & 63.55  & 69.97 & 71.68 (majority)  \\
\hline
Sentiue & {\bf 78.70 (1)}  & {\bf 79.35 (1)} & 71.68 (4)  \\
\hline
lsislif & 75.50 (3)  & 77.87 (3) & {\bf 85.84 (1)}  \\
\hline
EliXa (u) & 70.06(10) & 72.92 (7) &  79.65 (3) \\	
\hline  
EliXa (c) & 67.34 (14) & 71.55 (9) & 74.93 (5)  \\	
\hline	
\end{tabular}
}
\end{center}
\caption{Results obtained on the slot3 evaluation on restaurant data; ranking in brackets. }
\label{tab:slot3_res}
\end{table}

\section{Conclusions}
%
%

We have presented a modular and supervised ABSA platform developed to facilitate future experimentation in the field. We submitted runs corresponding to the slot2 and slot3 subtasks, obtaining competitive results. In particular, we obtained the best results in slot2 (OTE) and  for slot3 we obtain 3rd best result in the out-of-domain track, which is nice for a supervised system. Finally, a system for topic detection (slot1) is currently under development.

\section{Acknowledgments}
This work has been supported by the following projects: ADi project (Etortek grant No. IE-14-382), NewsReader (FP7-ICT 2011-8-316404), SKaTer (TIN2012-38584-C06-02) and Tacardi (TIN2012-38523-C02-01).

\bibliographystyle{naaclhlt2015}
\bibliography{references}

\end{document}